# Data-Driven Multi-step Demand Prediction for Ride-hailing Services Using Convolutional Neural Network

Chao Wang[1], Yi Hou[2], and Matthew Barth[1]

[1] University of California Riverside, Riverside, CA 92507, USA
[2] National Renewable Energy Laboratory, Golden, CO 80401, USA
cwang061@ucr.edu
Yi.Hou@nrel.gov
barth@ece.ucr.edu

**Abstract.** Ride-hailing services are growing rapidly and becoming one of the most disruptive technologies in the transportation realm. Accurate prediction of ride-hailing trip demand not only enables cities to better understand people's activity patterns, but also helps ride-hailing companies and drivers make informed decisions to reduce deadheading vehicle miles traveled, traffic congestion, and energy consumption. In this study, a convolutional neural network (CNN)-based deep learning model is proposed for multi-step ride-hailing demand prediction using the trip request data in Chengdu, China, offered by DiDi Chuxing. The CNN model is capable of accurately predicting the ride-hailing pick-up demand at each 1-km by 1-km zone in the city of Chengdu for every 10 minutes. Compared with another deep learning model based on long short-term memory, the CNN model is 30% faster for the training and predicting process. The proposed model can also be easily extended to make multi-step predictions, which would benefit the on-demand shared autonomous vehicles applications and fleet operators in terms of supply-demand rebalancing. The prediction error attenuation analysis shows that the accuracy stays acceptable as the model predicts more steps.

## 1 Introduction

In recent years, ride-hailing companies such as Uber, Lyft, DiDi Chuxing (China), and RideAustin have emerged as new and disruptive on-demand mobility services. In major cities, 21% of adults use ride-hailing services [1]. Accurate prediction of ride-hailing trip demand not only enables cities to better understand people's activity patterns, but also helps ride-hailing companies and drivers make informed decisions to reduce deadheading vehicle miles traveled, traffic congestion, and energy consumption. In the next wave of transportation innovations, such as mobility-as-a-service where on-demand shared automated vehicles transport people and goods in cities, accurate trip demand prediction will provide decision-making tools for automated vehicle fleet operators to optimize shared automated vehicle assignment holistically for the entire city.

Considerable interest in predicting the trip demand for taxi and ride-hailing trips has grown in the research community for the last a few years. Chang et al. [2] mined historical data to predict taxi demand distributions using clustering algorithms. Moreira-Matias et al. [3, 4] applied time series techniques to forecast taxi passenger demand. Gong et al. [5] proposed a machine learning model, XGBoost, to predict New York City taxi demand. Most of these efforts have focused on taxi trip demand, whereas the studies on ride-hailing demand prediction have been relatively limited. Recently, Ke et al. [6] introduced the fusion convolutional long short-term memory (LSTM) network (FCL-Net) to forecast passenger demand for on-demand ride services in Hangzhou, China, using real-world data provided by DiDi Chuxing. Wang et al. [7] developed a LSTM to predict the number of Uber pickups in New York City. Xu et al. [8] also developed an LSTM to predict taxi passenger demand for each small area in New York City. Liao et al. [9] conducted a thorough comparison between two deep neural network structures, deep spatio-temporal residual network (ST-ResNet) and FLC-Net, for taxi demand prediction using New York City taxi data. They found that deep neural networks outperform most traditional machine learning models when predicting taxi trip demand.

Most of the previous studies adopted a time-series model or recurrent neural network architecture, which are good at capturing time dependencies from historical data. One drawback of these models is that spatial information is usually lost in the modeling process. This paper proposes a convolutional neural network (CNN)-based deep learning model to predict the ride-hailing demand considering both temporal and spatial features. CNN is a class of deep and feed-forward artificial neural networks that is most commonly applied to analyzing visual imagery and proven to be a very efficient and well-performed image recognition algorithm [10, 11]. CNN was inspired by biological processes [12]. The connectivity pattern between neurons resembles the organization of the animal visual cortex. Individual cortical neurons respond to stimuli only in a restricted region of the visual field known as the receptive field. The receptive fields of different neurons partially overlap such that they cover the entire visual field. CNNs use a variation of multilayer perception designed to require minimal preprocessing [13]. Similarly, the convolutional layers of a CNN not only enable CNN to capture the local feature of the image data, but also reduce model parameters that need to be estimated. In recent years, CNNs have been utilized to solve transportation problems [14-17]. In this study, we uncover the possibility of learning the ride-hailing service demand patterns as images and make predictions by constructing the CNN input with proper spatial-temporal demand data.

Last year, DiDi Chuxing, the largest ride-hailing company in China, opened a new opportunity window for transportation researchers by sharing one month of trip data at city of Chengdu, China [18]. Thus, the proposed model was evaluated using DiDi Chuxing trip data. The remainder of the paper is organized as follows: the next section presents an overview of the data used for this research effort and formulates the problem to be solved in this paper; The third section discusses the methodology adopted for predictive analysis. The fourth section explains and presents the model results. The fifth and final section offers concluding thoughts and directions for future research.

## 2 Data and Problem Formulation

### 2.1 Data

The data used in this study are the ride-hailing trip request data in Chengdu, China offered by DiDi Chuxing [18]. DiDi Chuxing is the world's largest and most valuable ride-sharing behemoth, with a monopolistic investment and merger and acquisition portfolio in the ride and bike sharing industry across the globe [19]. As a leading mobile transportation platform, DiDi receives more than 25 million trip orders, collects more than 70 TB of new route data, processes more than 4,500 TB of data, and obtains more than 20 billion queries for route planning and 15 billion queries for geolocation. In 2017, DiDi announced its GAIA Initiative and decided to share a complete sample of the route and ride request data with the academic community. The dataset was collected from November 1 to November 30, 2016, at Chengdu, China. The request dataset contains 7 million ride request records with origin-destination (OD) points and the trip start and end times. The route dataset offers 1 billion data points of the trip global positioning system trajectories with a sampling rate of 2–4s. The rich, dense records enable us to pursue high-resolution demand prediction at a local level. The dataset mainly used for this study is the ride request records. The information includes the pick-up and drop-off location and time for each ride. Table 1 lists the details of the data description. Fig. 1 visualizes all the pick-up locations, which indicate the ride-hailing demand, over the observed month on a map of Chengdu.

The weather information is important for demand prediction considering the large impact of weather condition on the ride demand. We collected the open source hourly weather data from World Weather Online [20], including temperature, humidity, and weather conditions. Table 2 lists the parameters that we used in this study. For the convenience of model training, we assign each weather condition a numerical code.

**Table 1.** Ride request data from DiDi.

| Parameter | Sample | Comment |
|---|---|---|
| Order id | mjiwdgkqmonDFvCk3ntBpron5 mwfrqvI | Anonymized |
| Trip Start time | 1501581031 | Unix timestamp, in seconds |
| Trip end time | 1501582195 | Unix timestamp, in seconds |
| Pick-up longitude | 104.11225 | GCJ-02 Coordinate System |

| Parameter | Sample | Comment |
|---|---|---|
| Pick-up latitude | 30.66703 | GCJ-02 Coordinate System |
| Drop-off longitude | 104.07403 | GCJ-02 Coordinate System |
| Drop-off latitude | 30.6863 | GCJ-02 Coordinate System |

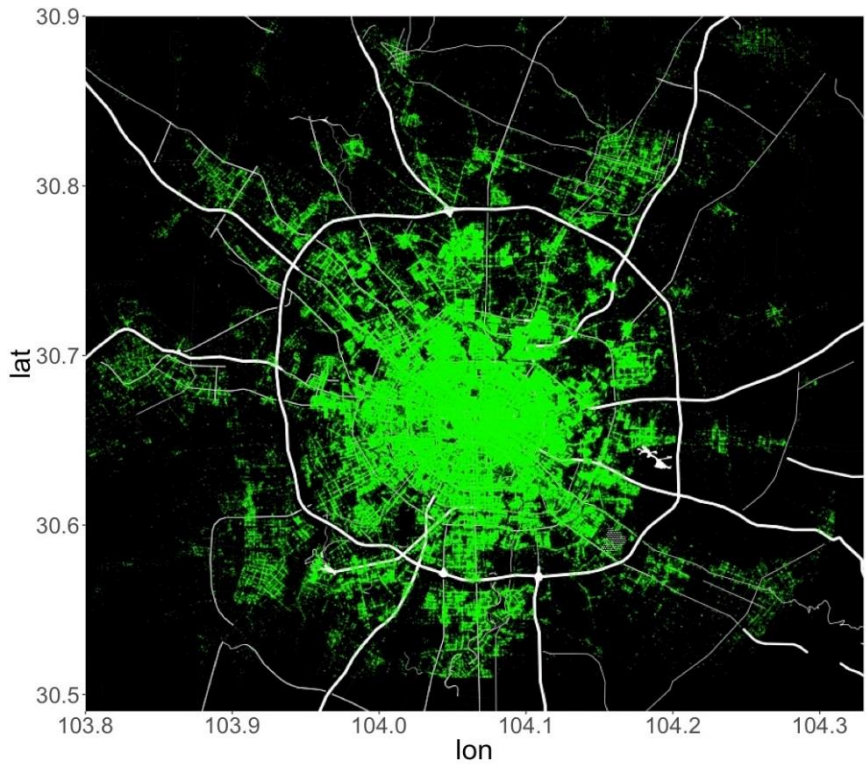

**Fig. 1.** Visualization of the pick-up locations in Chengdu, China.

**Table 2.** Weather data parameters and weather condition encoding.

| Parameter | Unit/Code | |
|---|---|---|
| Temperature | F° | |
| Humidity | % | |
| Weather condition | Unknown | 0 |
| | Clear | 1 |
| | Fog | 2 |
| | Haze | 3 |
| | Light Rain | 4 |
| | Light Rain Showers | 5 |
| | Mist | 6 |

| Parameter | Unit/Code | |
|---|---|---|
| | Mostly Cloudy | 7 |
| | Partly Cloudy | 8 |
| | Patches of Fog | 9 |
| | Scattered Clouds | 10 |

## 2.2    Problem Description

This study is a continuing work based on [7], in which the ride-hailing service demand prediction for New York City was proposed. We were able to predict the next 1-hour total number of demands for a certain region in the city using an LSTM network trained by previous demand in that region. The goal of this study is to improve the capability of the prediction method in terms of accuracy, computational efficiency, temporal granularity, and spatial scalability. Since the DiDi dataset offers denser ride request data, we can divide the city region into smaller zones for demand prediction while still having enough data points at each region for model training. As shown in Fig. 2, we select the urban core of the city and split it into $10 \times 10$ square grid cells. The longitude and latitude boundaries of the study region are also labeled. The high density of the ride-hailing demand also allows us to segment the prediction time interval in a smaller size and make more in-time prediction (e.g., predict the ride-hailing demand in the next 10 minutes). Another improvement we want to make is to involve the context information to the learning model, which will potentially increase the accuracy of the prediction. For instance, weather conditions heavily impact the ride-hailing demands, e.g., there might be more rides when it's raining or snowing. Predicting pick-up demand further into the future is always desirable. Therefore, instead of only predicting the pick-up demand in the next time step, the proposed model can predict the demand for multiple time intervals in the future.

Let $D_t = [d_{1_t}, d_{2_t}, \ldots d_{100_t}]$ be the number of demands for all 100 cells in time slot $t$, let $C_t = [min_t, day_t, t, temp_t, humi_t, wc_t]$ be the context-aware information of time $t$, where $temp_t, humi_t, wc_t$ represent the temperature, humidity, and weather conditions of time $t$, and $min_t, day_t, t$ are the minutes of the day, day of week, and global time step. Based on the above considerations, we defined the problem that needs to be solved as follows:

> ***Given*** $[D_{t-m}, D_{t-m+1}, \ldots, D_{t-1}]$ and $[C_{t-m}, C_{t-m+1}, \ldots, C_{t-1}]$ as input,
> ***provide*** a prediction model that outputs $D_{t+k}$,

where $m \geq 1$ is the number of past time steps used for prediction, $k \geq 0$ indicates the time step of ride-hailing demand that is predicted. In this study, $m$ is set to be 6 and $k \in [0,5]$, and the duration for each time slot is 10 minutes. Therefore, the past 60 minute ride-hailing demands and weather records are used to predict the next 10–60 minutes of ride-hailing demand.

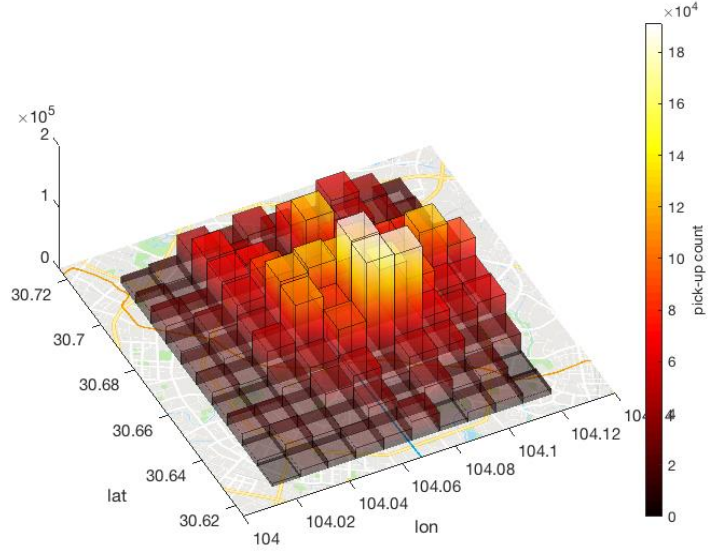

**Fig. 2.** Ride-hailing demand distribution at Chengdu.

## 3    Methodology

In this section, the methods to develop the context-aware multi-step ride-hailing demand prediction model are presented. Two deep learning methods are designed and tested for this study, the LSTM and the CNN. The LSTM methods are developed in author's previous work [7, 21] as a comparison. CNN is the major method that we focus on and explore in this paper.

Referring to the problem description section, the input of the model is $[D_{t-m}, D_{t-m+1}, ..., D_{t-1}]$ and $[C_{t-m}, C_{t-m+1}, ..., C_{t-1}]$, where $D_t$ is a $10 \times 10$ matrix representing the ride-hailing demand at time $t$ for all 100 zones. Inspired by computer vision and image processing techniques, $D_t$ can be treated as the frame of the image with 10 by 10 pixels. In order to involve the context-aware parameters, we added one more row of pixels to the frame and filled the first six pixels with the context-aware parameters $C_t = [min_t, day_t, t, temp_t, humi_t, wc_t]$. Thus, an $11 \times 10$ image input for the CNN is defined as shown in Fig. 3. The "look-back" parameter is $m = 6$, meaning six frames are input to the network at one time. So the input we constructed is a tensor, and its size is $6 \times 11 \times 10$, as shown in Fig. 4.

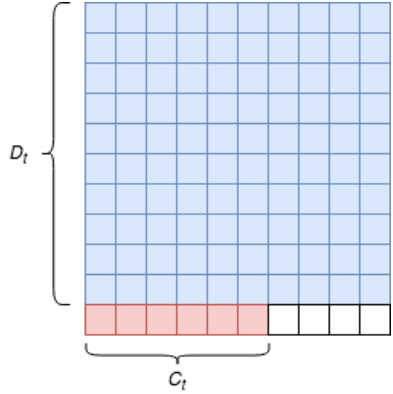

**Fig. 3.** CNN input construction.

The designed CNN structure for this study is shown in Fig. 4. The kernel size for the convolution and the max-pooling layer are labeled in the graph. The input is six frames of the matrices as we constructed above, representing that the model looks at the demand of the previous six time steps (60 min) and predicts the future time steps. There are two convolution layers, each followed by a max-pooling layer for down sampling. It is common to periodically insert a pooling layer between successive convolutional layers in a CNN architecture [22]. The pooling operation provides another form of translation invariance, operates independently on every depth slice of the input, and resizes it spatially [23]. The most common form of the pooling layer contains filters of size $2 \times 2$ with a stride of 2. It downsamples at every depth slice in the input by 2 along both the width and height, discarding 75% of the activations. The depth dimension remains unchanged. To increase the nonlinear properties of the decision function and of the overall network without affecting the receptive fields of the convolution layer, rectifier is used as activation function for the convolutional layer, also known as rectified linear unit (ReLU). The rectifier function gives an output $x$ if $x$ is positive and 0 otherwise:

$$f(x) = \max(0, x) \tag{1}$$

Following the two convolutional-pooling layer pairs is a flattening layer, which flattens the output from the last layer to a vector. The last two layers in the network are fully connected layers (256 hidden units) and the output layer. Since there are 10 × 10 regions in total for demand prediction, the output of the whole CNN is a 1 × 100 vector. For the multi-step prediction, we do not change the main characteristic of the CNN but only change the size of the output layer to output a 1 × 100$t$ vector, where $t$ is the number of steps predicted ahead. We applied zero padding to pad the input volume with zeros around the border while calculating the convolution to keep the output having same size with the input of the previous layer.

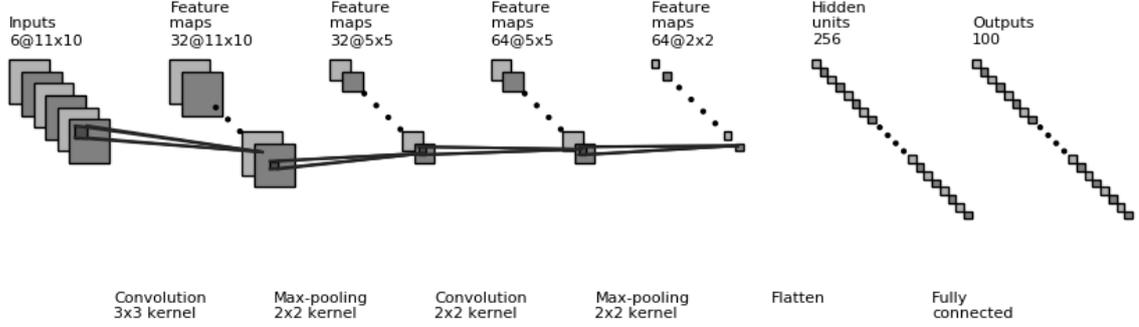

**Fig. 4.** CNN architecture design.

## 4 Results

In this section, we present the results and findings of the proposed CNN model. In addition to CNN, two other models are used for comparison. One is a trivial instanton model that uses the current time slot-observed ride-hailing demand as the prediction of the next time slot, another is a LSTM based deep learning model. The main design of the LSTM network for this study is similar to the author's previous work [7, 21] with a few modifications. Instead of using the one-dimensional inputs and outputs in the previous work, multi-dimensional inputs and outputs are used to enable including the context-aware information in the model. The modified model structure allows the LSTM network to predict the ride-hailing demand of all zones simultaneously. Fig. 5 visualizes the observation vs. the prediction of the three models. It shows that the predictions from the LSTM and CNN models are far better than the instanton model. For both LSTM and CNN, the correlation of the predicted demand and the observed demand is close to the ideal condition represented by the red line. More details of the prediction error are presented in Table 3. The error metrics that we used are weighted mean absolute percentage error (WMAPE) and mean absolute error (MAE). They are defined as follows:

$$WMAPE = 100\% \cdot \frac{\sum_{i=1}^{n}|y_i - \widehat{y_i}|}{\sum_{i=1}^{n}|y_i|} \tag{2}$$

$$MAE = \sum_{i=1}^{n}|y_i - \widehat{y_i}| \tag{3}$$

where $y_i$ is the observation, $\widehat{y_i}$ is the prediction, and $n$ is the number of samples. The WMAPE is a variant of mean absolute percentage error (MAPE). It is designed for measuring the percentage error but avoids problems where a series of small or zero denominators are present. Table 3 shows that CNN performs slightly better than LSTM in terms of error measures WMAPE and MAE.

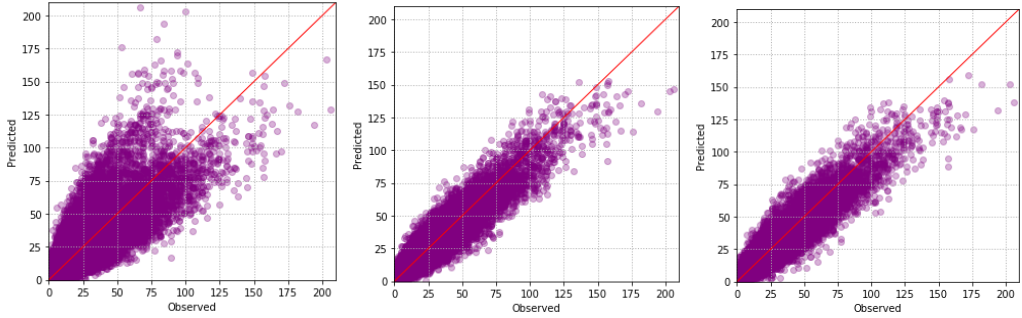

**Fig. 5.** Observation vs. prediction for next 10-minute time step (from left to right: instantons, LSTM, CNN)

**Table 3.** Model prediction error comparison (for next 10-minute time step).

| Model | Prediction Error | |
|---|---|---|
| | WMAPE | MAE |
| Instanton | 46.84 % | 4.6783 |
| LSTM | 24.97 % | 3.1078 |
| CNN | 23.59 % | 3.0616 |

The computational efficiency of LSTM and CNN is further examined. Table 4 lists the computing time for training and prediction for the two models. The comparison shows that despite more trainable parameters in the CNN model, the efficiency of the CNN model is over 30% faster for both the training and predicting when compared with LSTM.

**Table 4.** Model efficiency comparison.

| Model | Training Time | Prediction Time | # Net Parameters |
|---|---|---|---|
| LSTM | 17.59s | 0.2057s | 133,220 |
| CNN | 11.73s | 0.1407s | 172,452 |
| Improvement | 33.31% | 31.60% | |

For ride-hailing services, fleet operators need time to reassign vehicles to meet the future trip demand. Providing trip demand prediction further into the future is more useful than providing prediction of the immediate next time interval. Therefore, the accuracy of multi-step ride-hailing demand prediction was examined to test the long-term prediction capability of the proposed CNN model. The CNN architecture is slightly modified to accommodate 10, 20, 30, 40, 50, and 60 minutes ahead demand predictions. We only need to change the output size to fit the number of demands that are predicted for multiple steps, which is $100 \times 6$. Fig. 6 indicates that as the prediction becomes further ahead in the future, the prediction error of the CNN model is increase very slowly within an acceptable level. This indicates that the proposed CNN model performs well in long-term prediction. Fig. 7 shows the 60-minute ahead predictions vs. observations along time. It can be observed that the predictions align closely with the observed values.

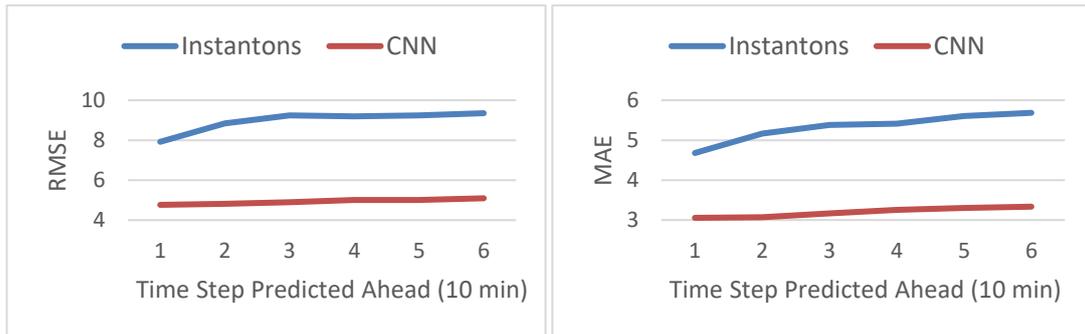

**Fig. 6.** Error attenuation with multi-step prediction.

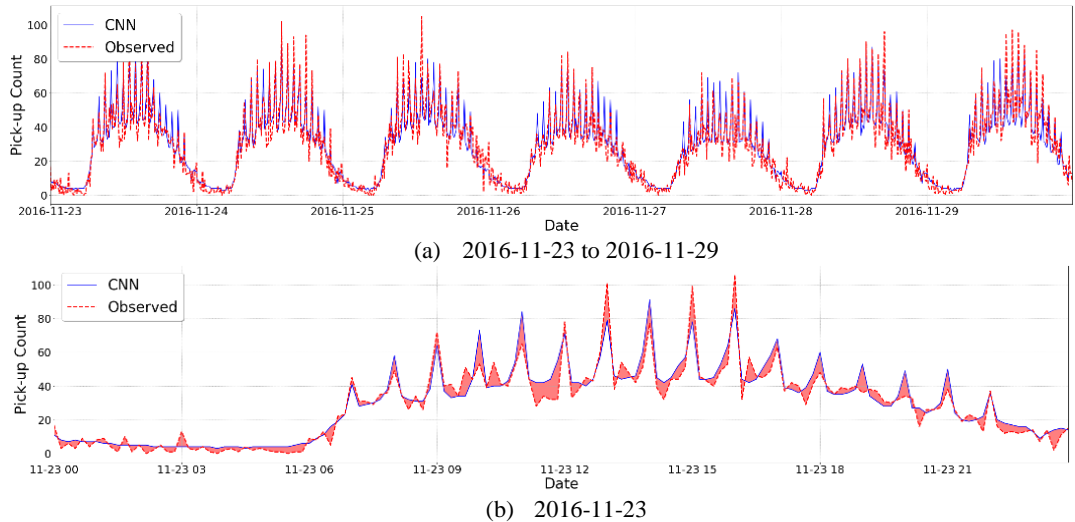

(a)  2016-11-23 to 2016-11-29

(b)  2016-11-23

**Fig. 7**. 60-minute-ahead ride-hailing demand predictions vs. observation (Zone #56, which is represented by the pixel at row 6, column 6, refer to Fig. 3).

## 5    Conclusion and Future Work

In this study, a CNN-based deep learning model is proposed for context-aware multi-step ride-hailing demand prediction. We utilized the 7-million trip records collected in

Chengdu, China, provided by DiDi Chuxing to train and test the model. The outcomes and findings are promising. We split Chengdu City into $10 \times 10$ square zones with 1-km side lengths. The CNN model can provide accurate demand predictions for all 100 zones every 10 minutes. The prediction accuracy significantly outperforms the baseline model and produces slightly lower error measures than the LSTM model, which the authors proved to effectively model ride demand in earlier published works. The computational efficiency of the CNN model is further examined. The result shows that although the number of trainable parameters in the CNN model are higher than in the LSTM model, the CNN model is 30% more computationally efficient for both training and predicting. The proposed model can also be easily extended for multi-step predictions that could be applied for on-demand shared automated vehicle operations. We found that the CNN model prediction accuracy is still satisfied when predicting ride-hailing demand 60 minutes ahead.

Predicting ride-hailing demand can benefit ride-hailing vehicle operation efficiency. For future work, a ride-hailing fleet dispatching system will be developed based on the demand prediction. Prior knowledge of the demand distribution around the city would help operators dispatching vehicles to the passengers' nearby locations before they make a ride request to provide more in-time service. Proactive fleet management will save vacant time and travel distance for the vehicles between rides.

## Acknowledgement

The authors want to thank DiDi Chuxing for providing the data for this study.

This work was authored in part by the National Renewable Energy Laboratory, operated by Alliance for Sustainable Energy, LLC, for the U.S. Department of Energy (DOE) under Contract No. DE-AC36-08GO28308. Funding provided by U.S. Department of Energy Office of Energy Efficiency and Renewable Energy Vehicle Technologies Office. The views expressed in the article do not necessarily represent the views of the DOE or the U.S. Government. The U.S. Government retains and the publisher, by accepting the article for publication, acknowledges that the U.S. Government retains a nonexclusive, paid-up, irrevocable, worldwide license to publish or reproduce the published form of this work, or allow others to do so, for U.S. Government purposes.